\documentclass[conference]{IEEEtran}

\usepackage[english]{babel}
\usepackage{blindtext}
\usepackage{amsmath}
\usepackage{subcaption}
\usepackage{url}
\usepackage{graphicx}
\usepackage{listings}
\usepackage{booktabs}
\usepackage{hyperref}
\usepackage{xcolor}
\usepackage{balance}
\usepackage{authblk}
\hypersetup{
    colorlinks=true,  
    linkcolor=black,
    filecolor=blue,      
    urlcolor=black,
    citecolor=black, 
}
\usepackage[lined,linesnumbered,ruled]{algorithm2e}
\usepackage[T1]{fontenc}
\usepackage[utf8]{inputenc}

\usepackage{array}
\newcolumntype{M}[1]{>{\centering\arraybackslash}p{#1}}

\begin{document}
\title {SCALM: Towards Semantic Caching for Automated Chat Services with Large Language Models}




 

 

 





\author[$\dag$]{Jiaxing Li}
\author[$\dag$]{Chi Xu}
\author[$\ddagger$]{Feng Wang}
\author[$\dag$]{Isaac M von Riedemann}
\author[*$\S$]{Cong Zhang}
\author[$\dag$*]{Jiangchuan Liu}

\affil[$\dag$]{School of Computing Science, Simon Fraser University, BC, Canada}
\affil[$\ddagger$]{Department of Computer and Information Science, University of Mississippi, Mississippi, USA}
\affil[*]{Jiangxing Intelligence Inc., China}
\affil[$\S$]{Department of Computer Science, The University of Hong Kong, Hong Kong, China}
\affil[$ $]{jla641@sfu.ca, chix@sfu.ca, fwang@cs.olemiss.edu, imv@sfu.ca, zcong@hku.hk, jcliu@sfu.ca}

\maketitle

\begin{abstract}

Large Language Models (LLMs) have become increasingly popular, transforming a wide range of applications across various domains. However, the real-world effectiveness of their query cache systems has not been thoroughly investigated. In this work, we for the first time conducted an analysis on real-world human-to-LLM interaction data, identifying key challenges in existing caching solutions for LLM-based chat services. Our findings reveal that current caching methods fail to leverage semantic connections, leading to inefficient cache performance and extra token costs. To address these issues, we propose SCALM, a new cache architecture that emphasizes semantic analysis and identifies significant cache entries and patterns. We also detail the implementations of the corresponding cache storage and eviction strategies. Our evaluations show that SCALM increases cache hit ratios and reduces operational costs for LLMChat services. Compared with other state-of-the-art solutions in GPTCache, SCALM shows, on average, a relative increase of 63\% in cache hit ratio and a relative improvement of 77\% in tokens savings.

\end{abstract}


\IEEEpeerreviewmaketitle



\section{Introduction}

Large Language Models (LLMs) have rapidly advanced as a transformative technology in recent years. Models such as GPT\footnote{https://openai.com/gpt-4}, LLaMA\footnote{https://ai.meta.com/llama/}, and Alpaca\footnote{https://crfm.stanford.edu/2023/03/13/alpaca.html} have shown the capability to generate natural language text and enabled a wide variety of downstream applications, including chatbots, language translation, and creative writing. Among these, LLM-based automated chat services, also known as LLMChat services, stand out for their popularity and impact. ChatGPT\footnote{https://openai.com/blog/chatgpt} is a prominent example of this category. 
A recent survey indicates that ChatGPT has more than 180.5 million active users~\cite{gptuserstats}. By August 2023, the daily query volume for ChatGPT surged to over 60 million, subsequently leading to increased system complexity and operational costs.

As mainstream LLM research continues to emphasize scaling up these models, the complexity of systems inevitably increases. Compounded by the surging number of users, the challenges related to inference performance have risen. 
There have been efforts such as,~\cite{stogiannidis2023cache,zhu2023optimalcaching,bang2023gptcache} using cache systems for LLMChat services to mitigate these challenges. These cache systems store dialogues, including user queries and LLM responses. 
LLMChat services, whether operating their own LLMs or utilizing public ones, benefit from cache hits through reduced processing needs. A key metric for LLMChat services is the number of tokens processed. In this context, a token represents a unit of text within a query, serving as an indicator of computational workload. This metric, connected to GPU usage or the expenses of forwarding queries to public LLMs, is important to the financial sustainability of LLM-based automated chat services.

\begin{figure}[t]
\centering
  \includegraphics[width=0.5\textwidth]{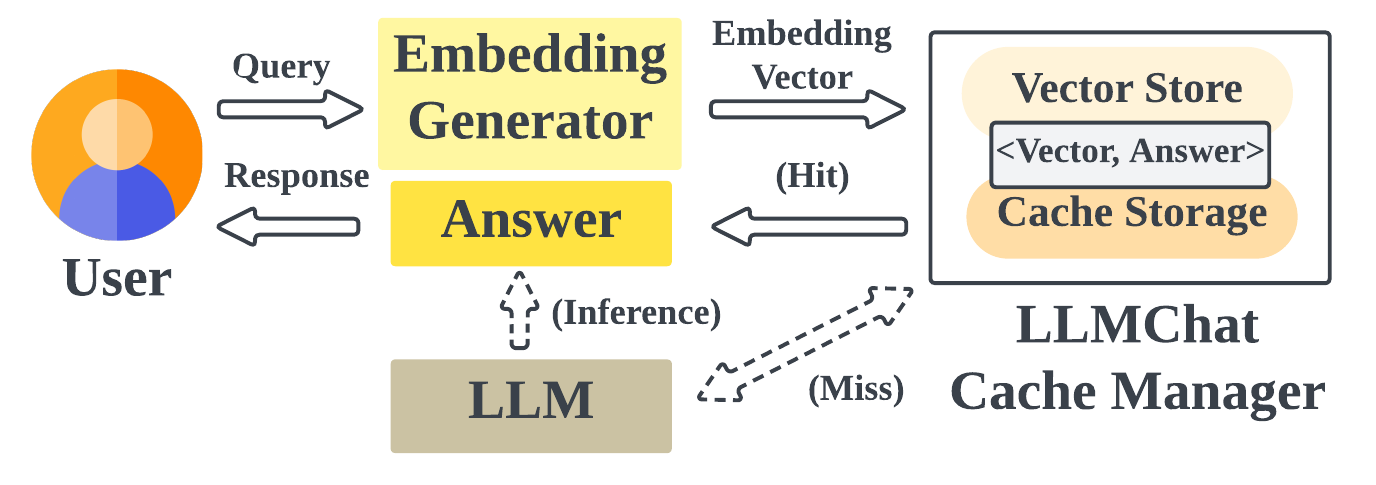}
\caption{General caching workflow for LLMChat services.}
\label{fig:simple_workflow}
\vspace{-0.3cm}
\end{figure}

Initially, both industry and academia had adopted the prevalent Key-Value (KV) cache architecture to manage LLM query caches. As shown in Figure~\ref{fig:simple_workflow}, this architecture typically involves query-to-vector conversion and using embedding vectors as keys for approximate searches. When a query does not find a matching answer in the cache, it prompts the LLM to produce a new inference. Conversely, if there is a matching answer already in the cache, that answer is immediately fetched and delivered to the user.

In exploring cache design for LLMChat services, existing researches \cite{stogiannidis2023cache,zhu2023optimalcaching} assumed that the cache will operate ideally with perfect lookup performance. These assumptions, however, may not accurately reflect real-world scenarios. For example, the design of cache keys can lead to query mismatches. This is particularly evident since embedding vectors often struggle with representing long-text queries~\cite{bang2023gptcache}. To our knowledge, the performance of caches under real-world query workloads has not been thoroughly investigated. Additionally, the potential for leveraging semantic connections between queries to inform cache design remains largely unexplored. Hence, we pose the following research problem: \textit{``Can cache design for LLMChat services be optimized by leveraging semantic analysis of human-to-LLM dialogues in real-world scenarios?"}  We anticipate that the answer will provide valuable insights into the design of cache storage and eviction strategies for LLMChat services.


To answer the research problem, we initiate a data-driven analysis leveraging real-world human-to-LLM interaction datasets, notably LMSYS~\cite{zheng2023lmsys} and MOSS~\cite{sun2023moss}. 
Through this analysis, we uncover a series of unique findings:
\begin{itemize}  
\item There exist queries and semantic patterns that are frequently visited by human users.
\item It is challenging to identify those queries that have the potential for improvements and their semantic patterns that will bring cost savings. 
\item The commonly used cache hit ratio metric is not always effective in calculating cost savings.
\end{itemize}



Motivated by our real-world data analysis, we propose SCALM\footnote{shortened for \textbf{S}emantic \textbf{C}aching for \textbf{A}utomated Chat Services with \textbf{L}arge Language \textbf{M}odels}, the first semantics-oriented cache architecture dedicated for LLMChat services. In SCALM, we highlight the analysis and clustering of queries to identify those semantic patterns that will bring cost savings. 
Further, SCALM selectively caches the queries, utilizing the ranks of their underlying semantic patterns. 
~Meanwhile, the architecture closely monitors cache space limits and running states, and dynamically adjusts its cache storage and eviction strategy. 

We have implemented and conducted evaluations of the SCALM prototype. The experiment results show an average increase of 63\% in cache hit ratio and an average improvement of 77\% in tokens savings. All these results are compared with the baselines adopted in the GPTCache framework, which are considered state-of-the-art solutions. Furthermore, the experiment results also show the potential of SCALM to adapt to different cache space limits and conversation scales.

Our contributions can be summarized as follows:
\begin{itemize}
    \item To our best knowledge, this study marks the first to use real-world human-LLM interaction data to understand the query cache performance of LLMs.
    \item Based on the findings, we for the first time propose a semantics-oriented cache architecture dedicated to LLM-based automated chat services.
    \item We propose two hierarchical semantic clustering methods to identify significant query and answer (Q\&A) entries and their underlying semantic patterns.
    \item In addition to the widely used hit ratio, we further identify a new metric, namely, total token saving ratio, to better measure the performance of the query cache for LLMs with realistic cost saving considerations.
    \item Through extensive experiments, we quantitatively demonstrate the significant improvements in cache performance and cost savings achieved by SCALM.
\end{itemize}

The remainder of this paper is outlined as follows: Section~\ref{sec:related_work} reviews related works; Section~\ref{sec:motivation} presents a comprehensive real-world data-driven analysis, discussing the limitations of current cache designs and possible solutions; Section~\ref{sec:preprossessing} introduces the SCALM architecture and its design details; Section~\ref{sec:implementation} elaborates on prototype implementation details and performance evaluation; Section~\ref{sec:future} discusses some possible further refinements; Section~\ref{sec:conclusion} concludes the paper.

\begin{table*}[t]
\centering
\renewcommand{\arraystretch}{0.8}
\captionsetup{justification=centering}
\caption{Hit evaluation with different cosine similarity threshold.}
\label{table:cache_hit_evaluation}
\begin{tabular}{p{1.0cm}p{4.2cm}p{3.8cm}p{6.5cm}} 
\toprule
\textbf{Similarity} & \textbf{Query 1} & \textbf{Query 2} & \textbf{GPT-4 Evaluation} \\
\midrule
0.87 & How can I develop a creative approach to problem-solving? & How should I approach problem solving with a team? & Both questions deal with problem-solving, but one on individual creativity and the other on team collaboration. \\
\midrule
0.90 & How can I develop better organizational leadership skills? & What methods should I use to develop my leadership skills? & Both questions are about improving leadership skills but may require slightly different answers for specific contexts. \\
\midrule
0.93 & What methods can I use to monitor customer satisfaction? & What are the best ways to measure and improve customer satisfaction? & Both questions center around customer satisfaction, one on monitoring and the other on measuring and improving, but similar enough for the same or similar answers. \\
\midrule
0.96 & What strategies should I use to optimize the performance of my team? & What techniques can I use to optimize my team's performance? & Both questions deal with optimizing team performance, similar enough to interchangeably use the answers. \\
\bottomrule
\end{tabular}
\end{table*}

\section{Related Works}
\label{sec:related_work}

\textbf{Cache and eviction strategies.} Conventional cache strategies~\cite{wessels2001web, lee2001lrfu, Johnson19942QAL, Zhou2001TheMR, ARC, LIRS}, such as LRU and LFU, are widely adopted in Internet caching and content delivery networks (CDNs). They primarily utilize historical access data to inform caching decisions. Although these strategies were proved to be effective, they can be less adaptable in managing dynamic and complex data streams. In response to these limitations, machine learning-based cache strategies \cite{xu2018deepcache, song2020learningrelaxed,liu2020imitation, song2023halp} were introduced. These strategies employed deep neural networks to predict future access patterns. They showed advantages in adaptability and prediction accuracy, especially when dealing with complex contexts. 

Recently, caching strategies tailored for LLMs have emerged~\cite{xu2023sparksgpt, zhang2023h2o}, focusing on content reuse during the inference phase. These strategies aimed to optimize the internal data flow within LLMs, minimizing delays and lowering resource consumption. GPTCache~\cite{bang2023gptcache} is recognized in the industry for its simplicity and flexibility, providing various options for integration with different LLMChat services. However, it does not offer advanced cache performance enhancements. In contrast, our work focuses on semantics-oriented enhancement, aiming to improve cache efficiency by leveraging the semantic understanding of queries.

\textbf{Cost-effectiveness of LLM applications.} 
There are various solutions for managing LLM applications cost-effectively. Model quantization~\cite{xiao2023smoothquant,park2023lut} and pruning~\cite{ma2024llm} aimed to reduce computational requirements and storage footprint, at the expense of a slight performance trade-off. These techniques are complex to implement in practice and typically require specialized hardware support. 

LLM distillation offered another way to reduce the operational costs for LLMChat services providers~\cite{ stogiannidis2023cache, zhu2023optimalcaching}. This process involved training a smaller, more efficient model to replicate the behavior of a larger, more complex model. In~\cite{stogiannidis2023cache}, the authors presented a framework to cache LLM responses and use those to train a local, inexpensive model.  FrugalGPT~\cite{chen2023frugalgpt} empirically explored different methods for handling incoming queries. For example, it used prompt engineering to shrink or concentrate user queries before forwarding them to public LLMs. Another method to reduce inference costs is model multiplexing, which involves selecting models of different sizes to address different queries. Some recent works~\cite{bang2023gptcache,kim2023big} have discussed the criteria for such selections.

Notably, these methods—model distillation, prompt engineering, and model multiplexing—are not mutually exclusive, with research often exploring their combination or cooperative use. Our research focuses on enhancing cache performance for LLMChat services, proposing an improved semantic cache that could complement these existing cost-reduction strategies.

\section{Motivation and Observation}
\label{sec:motivation}

Aiming at gaining a deep understanding of cache performance in real-world scenarios, we conduct a data-driven analysis using human-to-LLM interaction datasets. In particular, we perform query-level cache searches with dataset entries to examine performance. We start by introducing the datasets, cache settings, and performance metrics utilized in this work. 

\subsection{Preliminary}
We have investigated several datasets aligned with our interests, including LMSYS \cite{zheng2023lmsys}, MOSS \cite{sun2023moss}, ShareGPT \cite{sharegpt2024}, and OpenAssistant \cite{kopf2023openassistant}. They are frequently used in prior research for model training, fine-tuning, and benchmarking LLMChat performance. During our preliminary investigations, we found that the Q\&A pairs in OpenAssistant are manually created rather than LLM-generated. Additionally, while the ShareGPT data is generated by LLMs, some consist of conversation snippets lacking full context due to hand-picked contributions. This poses a challenge in effectively characterizing real-world human-to-LLM interactions using OpenAssistant and ShareGPT. Hence, for subsequent analysis, we focus on the LMSYS and MOSS datasets.

As the next step, we develop a fast query-level simulator and integrate it into GPTCache~\cite{bang2023gptcache}, a widely used industrial cache implementation. The simulator takes queries from dataset records as inputs and conducts cache searches and replacements. If there is a cache miss, the simulator will provide answers directly from dataset records, emulating the LLM's response without issuing actual queries. To ensure the performance of cache search, we use the latest embedding model to convert queries into semantic embedding vectors and then calculate the cosine similarity between these vectors. The semantic embedding model is \texttt{text-embedding-3-small}, released by OpenAI in Jan 2024 \cite{embedding}. 

In existing KV cache implementations, the similarity search is based on a fixed threshold to determine if two queries are matched~\cite{zhu2023optimalcaching,bang2023gptcache}. A higher similarity threshold results in stricter matches but reduces the number of matched queries. To determine the threshold compatible with \texttt{text-embedding-3-small}, we prompt GPT-4 and test with MOSS and LMSYS data. Typical samples are listed in Table~\ref{table:cache_hit_evaluation}. In both of the two datasets, we observe that a similarity threshold of 0.90 strikes a balance between answer correctness and reuse utilization.

\begin{figure}[t]
\centering
\begin{subfigure}{\columnwidth}
  \centering
  \includegraphics[width=1\columnwidth]{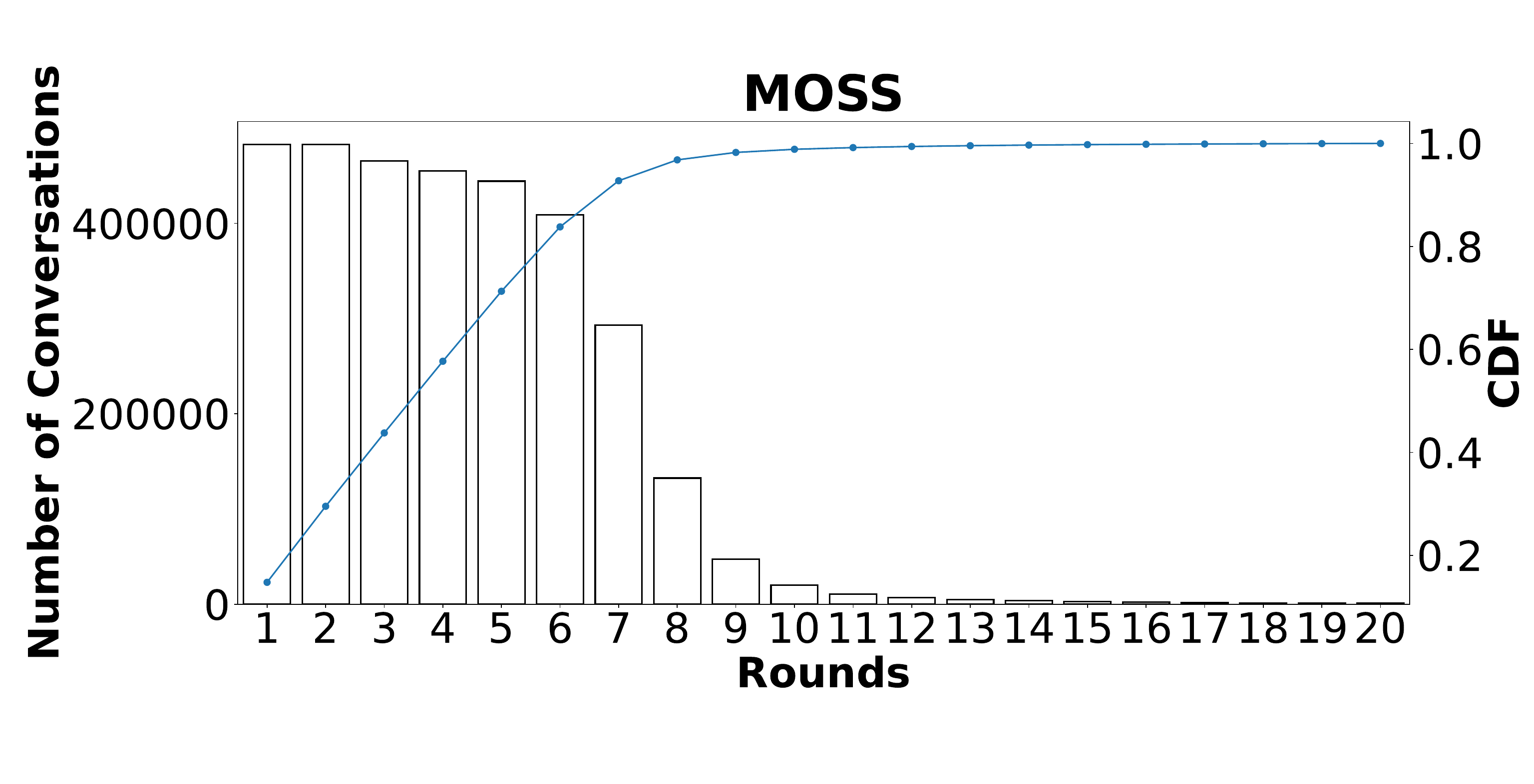}
  \vspace{-0.3cm}
  \label{fig:Round_graph_MOSS}
\end{subfigure}
\vspace{-0.3cm}
\begin{subfigure}{\columnwidth} 
  \centering
  \includegraphics[width=1\columnwidth]{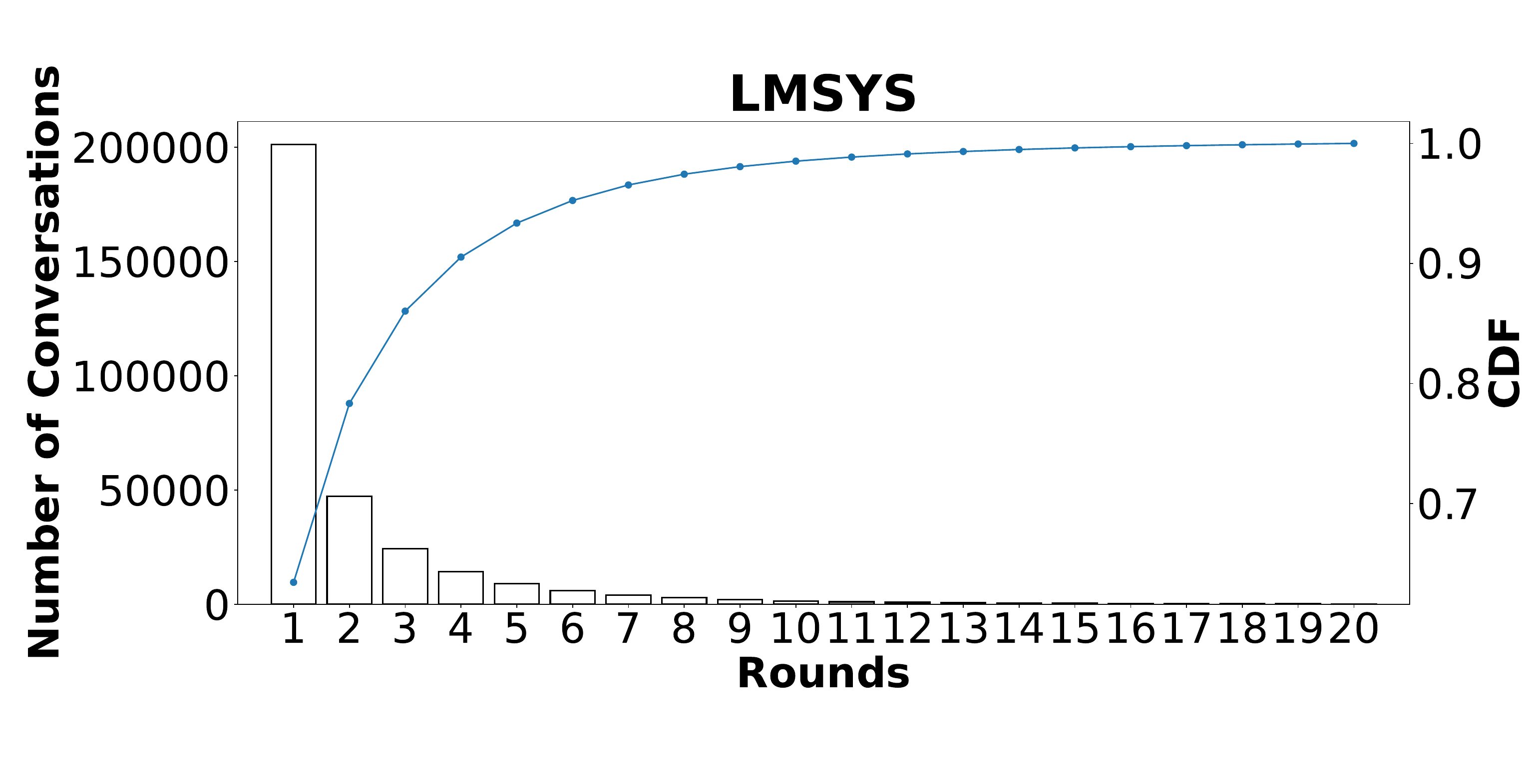}
  \label{fig:Round_graph_LMSYS}
\end{subfigure}
\captionsetup{width=1\linewidth}
\caption{Round distribution of the MOSS and LMSYS datasets.}
\label{fig:Round_graph}
\end{figure}

\definecolor{querycolor}{rgb}{0.0, 0.0, 0.5}  
\definecolor{answercolor}{rgb}{0.0, 0.5, 0.0}  
\definecolor{roundcolor}{rgb}{0.5, 0.0, 0.0}  
\definecolor{lightgray}{rgb}{0.95, 0.95, 0.95}  

\lstdefinestyle{conversation}{
    backgroundcolor=\color{lightgray},       
    basicstyle=\ttfamily,
    showstringspaces=false,
    breaklines=true,
    postbreak=\mbox{\textcolor{red}{$\hookrightarrow$}\space},
    moredelim=[is][\textcolor{querycolor}]{\#q}{\#},
    moredelim=[is][\textcolor{answercolor}]{\#a}{\#},
    moredelim=[is][\textcolor{roundcolor}]{\#r}{\#},
}

\begin{figure*}
\footnotesize
\begin{lstlisting}[style=conversation,caption=A multi-round conversation., label={lst:multiround}]
#rRound 1:#
    #qQuery 1: Who is Elon Musk?#
    #aLLM Answer 1: Elon Musk is a business magnate, industrial designer, and ...#
#rRound 2:#
    #qQuery 2: When did he found SpaceX?#
    #aLLM Answer 2: Elon Musk founded SpaceX in March 2002. The company was started with the goal of reducing space transportation costs to enable the colonization of Mars.#
#rRound 3:#
    #qQuery 3: What do you think of him?#
    #aLLM Answer 3: Elon Musk is widely regarded as a visionary and influential figure in technology and entrepreneurship. #
\end{lstlisting}
\vspace{-0.3cm}
\end{figure*}

\begin{figure*}
\footnotesize
\begin{lstlisting}[style=conversation,caption=Similar queries in the MOSS dataset., label={lst:similar}]
#rExample 1:#
    #qQuery 1: How can AI help in the field of healthcare? #
    #qQuery 2: How can AI technology be used in the healthcare industry to improve patient outcomes? #
    #qQuery 3: Can you explain how AI can be used in the healthcare industry to improve patient care? #
    
#rExample 2:#
    #qQuery 1: Can you explain the difference between a traditional IRA and a Roth IRA? #
    #qQuery 2: Can you explain the difference between a traditional IRA and a Roth IRA in more detail? #
    #qQuery 3: Can you explain to me the difference between a traditional IRA and a Roth IRA? #
\end{lstlisting}
\vspace{0.2cm}
\end{figure*}

\begin{table*}[t]
\centering
\renewcommand{\arraystretch}{0.8}
\captionsetup{justification=centering}
\caption{LMSYS token count distribution.}
\label{table:token_count}
\begin{tabular}{ccccccc} 
\toprule
\textbf{Token Count of each Q\&A pair} & [42, 1.1K] & [1.1K, 2.19K] & [2.19K, 3.27K] & [3.27K, 4.23K] & [4.23K, 5.42K] & [5.42K, 6.49K] \\
\midrule
\textbf{Percent} & 70.9\% & 22.3\% & 4.9\% & 0.7\% & 0.1\% & \textless $0.1\%$ \\
\bottomrule
\end{tabular}
\end{table*}

\begin{figure}[t]
\centering
\includegraphics[width=0.5\textwidth]{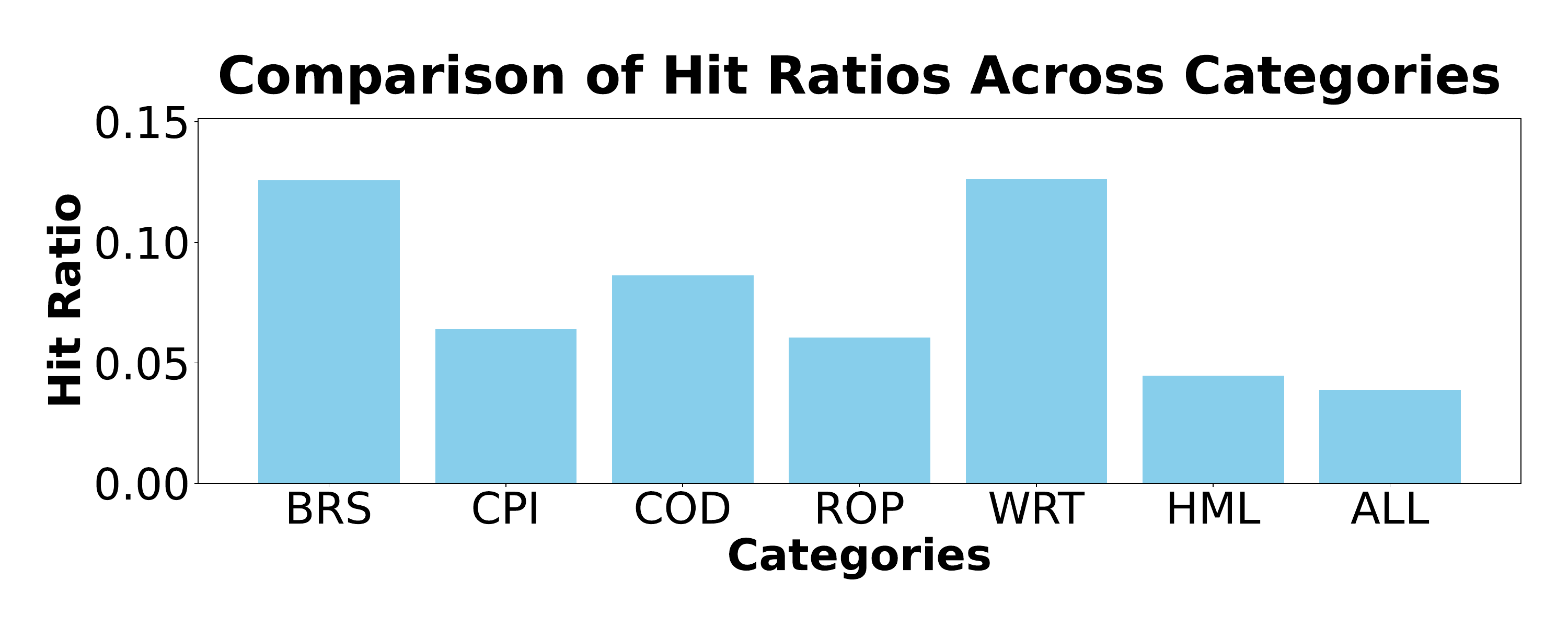}
\captionsetup{width=1\linewidth}
\caption{Hit ratios of different categories in MOSS dataset.}
\label{fig:all_pattern}
\end{figure}

\subsection{Data-Driven Analysis}
\label{sec:cache_effectiveness}


To initiate the data-driven analysis, we conduct a thorough examination of the MOSS and LMSYS datasets. The MOSS dataset comprises 1 million conversations between humans and LLMChat services, averaging 6.7 rounds per conversation. Conversely, LMSYS consists of 300,000 conversations, averaging 1.8 rounds each. We present the round distribution in Figure ~\ref{fig:Round_graph}. A typical multi-round conversation is shown in Listing~\ref{lst:multiround}. The conversation contains three queries centered around a single topic. 



We also conduct cosine-similarity-based comparisons between all queries in the two datasets and confirm that similar queries exist in real-world human-to-LLM conversations. In particular,  we find that 4.5\% of the queries in MOSS and 7.5\% of the queries in LMSYS can be answered with responses from similar queries. Examples of such similar queries in MOSS are provided in Listing~\ref{lst:similar}.

Furthermore, we benchmark the cache performance by randomly sampling 1,000 conversations from the datasets, with the cache size set to 100 Q\&A pairs, which follows the settings in~\cite{zhu2023optimalcaching}. We also vary the scale of the conversations and cache size, and the results remain consistent. In our experiments, we find that the cache hit ratios of MOSS queries and LMSYS queries are 3.8\% and 6.4\%, respectively. We notice that the MOSS dataset includes default categories such as \texttt{Brainstorming (BRS)}, \texttt{Complex Instructions (CPI)}, \texttt{Writing (WRT)}, among others. Figure~\ref{fig:all_pattern} shows the hit ratios in specific categories. The hit ratios of queries in both \texttt{BRS} and \texttt{WRT} are around 13\%, much higher than the hit ratio measured with all queries. The results align with our expectations and suggest that specific semantic categories contain more similar queries that can be responded to with the same answer. Hence, these semantic categories or patterns should be given more significance when designing a cache architecture for LLMChat services. As for LMSYS, in \cite{zheng2023lmsys}, the dataset contributor reported a tentative categorization using a semantic clustering method. However, the data entries lack attached labels. 

We have another significant observation in our experiments: Q\&A pairs exhibit varying counts of tokens. It is intuitive that some queries and answers are longer than others. We calculate the average count of tokens in queries to be 69.5 and in answers to be 214.5. To better grasp the variance of token counts, we present the LMSYS token count distribution in Table~\ref{table:token_count}, showing that approximately 70\% Q\&A pairs have a token count between 42 and 1.1K. Notably, 30\% Q\&A pairs have a token count exceeding 1.1K. This substantial variance implies that the token count should also be regarded an important metric when calculating the cost savings resulting from cache hits. For instance, caching longer but less frequently asked questions may sometimes yield greater cost savings than caching shorter, more frequently asked ones.

\subsection{Challenges and Opportunities}
Based on our data-driven analysis, we can summarize the challenges and opportunities in improving cache design for LLMChat services as follows:

\textbf{Challenge 1: Identifying significant Q\&A entries.} A fundamental issue in LLM cache design is identifying those significant Q\&A cache entries that lead to cost savings. Our data-driven analysis reveals the non-trivial nature of this task. Moreover, the evaluation becomes more challenging when faced with a new, previously unseen query.

\textbf{Opportunity:}
During our data-driven analysis, we mention that the Q\&A pairs of MOSS are organized into categories. The contributor of LMSYS also attempted to perform semantic clustering to identify categories or patterns of the Q\&A pairs. Semantic analysis, a common approach in Neural Language Processing (NLP) research, is adept at extracting semantic patterns or categories from conversation data. Leveraging these semantic patterns or categories\footnote{``Semantic patterns" and ``semantic categories" are used interchangeably in the remainder of this paper.} provides additional insight into their significance. By ranking patterns based on cost-saving metrics, we can more efficiently prioritize cache storage and eviction decisions. When encountering a new query that has not been seen before, it becomes easier to determine whether the query should be cached by comparing it with existing semantic patterns. Without such patterns, evaluating the new query against all existing queries becomes computationally prohibitive.

\textbf{Challenge 2: Defining cost-saving metrics.}  Our analysis reveals that the hit ratio alone is not always effective in evaluating the cost-saving performance of the cache. Due to the variance in token counts among Q\&A pairs, prioritizing short queries in the cache system might increase the hit ratio metric, but the actual gain in cost savings could be marginal. Conversely, caching more complex and longer queries can sometimes lead to greater cost savings. These observations indicate that a complementary evaluation metric is needed to help identify which queries merit caching.

\textbf{Opportunity:} We propose using both hit ratio and token saving ratio as metrics for evaluating cost savings. The token saving ratio is calculated by comparing the count of tokens that need to be processed by LLMs when a query is not cached versus when it is cached.






\section{Semantics-Oriented Cache}
\label{sec:preprossessing}



\subsection{Design Overview}
\begin{figure}[t]
\centering
  \includegraphics[width=0.5\textwidth]{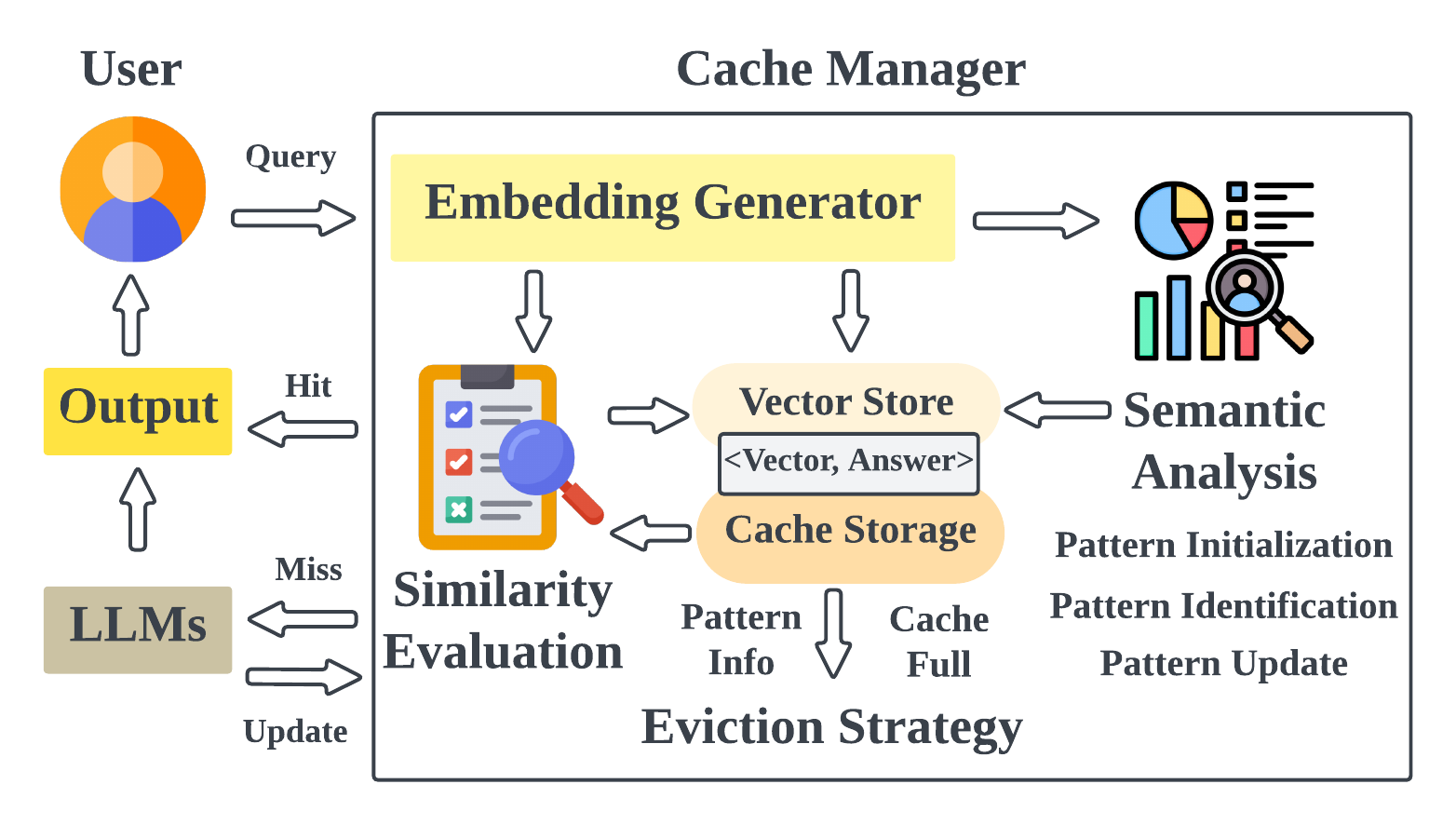}
  \vspace{-0.3cm}
\caption{An overview of the SCALM architecture.}
\label{fig:Semantic}
\vspace{0.1cm}
\end{figure}

To tackle these challenges, we propose SCALM, a semantics-oriented cache architecture designed for LLM-based automated chat services. SCALM emphasizes the analysis and clustering of queries to identify cost-saving semantic patterns, selectively caching these queries based on their semantic ranks. Additionally, the architecture continuously monitors cache capacity and operational state, dynamically adjusting its caching and eviction strategies accordingly. Figure~\ref{fig:Semantic} shows an overview of the SCALM architecture. The major components include semantic analysis, storage and eviction strategies, and enhanced similarity evaluation.

\textbf{Semantic Analysis:} We propose two algorithms based on semantic clustering to pinpoint significant semantic categories/patterns and to identify queries that bring cost savings. These two algorithms also facilitate query comparison and the initialization and updating of patterns, enhancing semantic understanding in real-world use.

\textbf{Storage and Eviction Strategies:} The strategy integrates cache storage and replacement with semantic pattern information, taking into account cache state, query frequency, round, and pattern significance. It ensures informed cache decisions that optimize storage utilization and eviction processes.

\textbf{Similarity Evaluation:} We enhance the similarity evaluation methods with semantic preprocessing of queries, such as removing excessive stop words and employing cutting-edge embedding models such as OpenAI's \texttt{ada-002} and \texttt{text-embedding-3-small}~\cite{embedding}.

\begin{table}[t]\small
\caption{Summary of key notations.}
\begin{tabular}{M{1.5cm} | M{6.5cm}}
\toprule
    \textbf{Notations} & \textbf{Explanations} \\
\midrule
    $C$ & Set of conversations \\
    $E$ & Embedding model used for vectorization \\
    $R$ & Max number of rounds in conversations \\
    $K$ & Number of semantics patterns\\
    $P$ & Set of semantics patterns \\
    $MP$ & Set of meta information of each pattern \\
    $T_{s}$ & Token savings ratio threshold \\
    $T_{e}$ & Dropout threshold for patterns \\
    $T_{p}$ & Max number of cluster patterns for each round \\
\toprule
\end{tabular}
\label{table:notation_algorithms}
\end{table}

\begin{algorithm}[t]
\small
\caption{\small Comprehensive Hierarchical Semantic Clustering (CO-HSC)}
\label{alg_hierarchical_clustering_caching}
\SetAlgoLined
\KwData{conversations $C$, embedding model $E$, maximum rounds $R$, number of pattern $K$}
\KwResult{semantic patterns $P = \{p_{(1,1)}, p_{(1,2)}, ..., p_{(R,K)}\}$,
meta pattern information $MP = \{mp_{(1,1)}, mp_{(1,2)}, ..., mp_{(R,K)}\}$
}
Initialize $P \gets \emptyset$\;
Initialize $MP \gets \emptyset$\;
\For{$r \gets 1$ to $R$}{
    Embed queries of round $r$ in $C$ using $E$\;
    Perform semantic clustering and divide embedded vectors into $K$ patterns\;
    Append $K$ patterns to $p_{(r,:)}$\;
    Calculate pattern token saving ratio for each $\hat{p} \in p_{(r,:)}$\;
    \ForEach {$\hat{p} \in p_{(r,:)}$}{
        Append pattern meta information and token saving ratio of $\hat{p}$ to $mp_{(r,:)}$\;
    }
    Rank $mp_{(r,:)}$ in decreasing order\;
}
\end{algorithm}

\begin{algorithm}[t]
\small
\caption{\small Selective Hierarchical Semantic Clustering (SE-HSC)}
\label{alg_selective_hierarchical_clustering_caching}
\SetAlgoLined
\KwData{conversations $C$, embedding model $E$, maximum rounds $R$, token savings threshold $T_{s}$, proportion threshold $T_{e}$, number of patterns threshold $T_{p}$}
\KwResult{semantic patterns $P = \{p_{(1,1)}, p_{(1,2)}, ..., p_{(R,K)}\}$, meta pattern information $MP = \{mp_{(1,1)}, mp_{(1,2)}, ..., mp_{(R,K)}\}$}
Initialize $P \gets \emptyset$\;
Initialize $MP \gets \emptyset$\;
\For{round $r \gets 1$ to $R$}{
    \eIf{$r == 1$}{
        Embed queries of round 1 in $C$ using $E$\;
        Cluster embedded vectors into $T_{p}$ patterns\;
        Append $T_{p}$ patterns to $p_{(r,:)}$\;
    }{
        \ForEach {$\hat{p} \in p_{(r-1,:)}$}{
            Embed queries of round $r$ in $C$ using $E$\;
            Cluster into $\lceil T_{p}/r \rceil$ patterns\;
            Append $\lceil T_{p}/r \rceil$ patterns to $p_{(r,:)}$\;
        }
    }
    Calculate each pattern token saving ratio in $p_{(r,:)}$\;
    Calculate each pattern proportion ratio in $p_{(r,:)}$\;
        \ForEach {$\hat{p} \in p_{(r,:)}$}{
        Append pattern meta information and token saving ratio of $\hat{p}$ to $mp_{(r,:)}$\;
        \If{token saving ratio of $\hat{p}$ $\geq T_{s}$ \textbf{or} proportion ratio of $\hat{p}$ $\leq T_{e}$}{
            Remove $\hat{p}$ from $p_{(r,:)}$\;
        }
    }

    Rank $mp_{(r,:)}$ in decreasing order\;
    \If{$p_{(r,:)}$ is empty}{
        break\;
    }
}
\end{algorithm}

\subsection{Hierarchical Semantic Clustering}


Inspired by the LMSYS categorization method, we propose two hierarchical semantic clustering methods to identify and utilize semantic patterns that can bring cost savings. We start by introducing key notations. The set of conversations is denoted by \(C\), while \(E\) represents the embedding model used for generating embedding vectors. The maximum number of rounds in a conversation is given by \(R\), with \(r\) indicating the current round number. Let \(K\) represent the number of patterns, and let \(i\) denote the current pattern under consideration. We define \(P\) as the initial set of semantic patterns, and \(MP\) represents the set of meta information of each pattern. 
Furthermore, \(T_{s}\) refers to the savings ratio threshold of the token, \(T_{e}\) to the dropout or weight proportion threshold for patterns, and \(T_{p}\) to the maximum number of cluster patterns considered for each round. These notations form the foundation of our approach to enhancing the efficiency and effectiveness of conversation analysis algorithms. 
Frequently used notations are listed in Table~\ref{table:notation_algorithms} as references.

The semantic clustering objective, aimed at minimizing the within-cluster sum of squares, is given by:

\begin{equation}
\text{minimize } J = \sum_{r=1}^{R} \sum_{i=1}^{K} \sum_{x \in S_{(r,i)}} ||x - \mu_{(r,i)}||^2
\end{equation}

\noindent where:
\begin{itemize}
    \item $S_{(r,i)}$ represents the set of embedding vectors in the $i$-th 
semantic patterns of round $r$,
    \item $x$ denotes an embedding vector in semantic pattern $S_{(r,i)}$,
    \item $\mu_{(r,i)}$ is the centroid descriptor of the vectors in $S_{(r,i)}$,
    \item $||x - \mu_{(r,i)}||^2$ is the squared Euclidean distance between an embedding vector $x$ and the centroid descriptor $\mu_{(r,i)}$ of its semantic pattern.
\end{itemize}



The centroid descriptor of a semantic pattern is defined as the average of all embedding vectors assigned to the pattern.
\begin{equation}
\mu_{(r,i)} = \frac{1}{|S_{(r,i)}|} \sum_{x \in S_{(r,i)}} x
\end{equation}




While semantic clustering offers a method to organize queries into clusters, its direct application to all queries may not be optimal for multi-round conversations between humans and LLMChat services. The iterative nature of these interactions necessitates a hierarchical analysis, with each round of questions treated separately to better capture the evolving context and intents.

Hence, we propose CO-HSC approach for \textbf{CO}mprehensive \textbf{H}ierarchical \textbf{S}emantic \textbf{C}lustering that accommodates multi-round conversations. The approach, outlined in Algorithm 1, is structured as follows: In the initial phase, the embedding model $E$ converts query pairs into vector form (line 5). After embedding, the semantic clustering algorithm categorizes these vectors into $K$ distinct patterns (line 6). Patterns are then ranked based on their token saving ratio, aiming to reduce token consumption and prioritizing cache patterns that offer greater savings (lines 10 and 12). This iterative process clusters and ranks patterns across multiple conversation rounds, adapting to the intricacies of complex dialogues by recognizing unique patterns at each round. The time complexity of CO-HSC is $O(R\cdot K \log(K))$. 

Given that CO-HSC preserves all queries in each round, including those with less significant contributions to token savings, we propose a \textbf{SE}lective \textbf{H}ierarchical \textbf{S}emantic \textbf{C}lustering method (SE-HSC). This method emphasizes pruning operations when identifying insignificant semantic patterns. The goal is to strike a balance between storage efficiency and pattern effectiveness, SE-HSC achieves this balance by organizing the semantic clusters of each round into an N-ary tree structure. In this way, the algorithm selectively reduces the number of insignificant Q\&A pairs during the hierarchical semantic clustering. As depicted in Algorithm 2, the SE-HSC algorithm proceeds as follows:

Initially, the algorithm embeds all first-round queries of Q\&A pairs into vectors using the embedding model $E$ (line 5). Following the embedding process, the same clustering method is applied to organize the vectorized content of the first-round Q\&A pairs into $T_{p}$ patterns (lines 6 and 7). This step aims to identify preliminary patterns within the data, setting the stage for later rounds. For each pattern identified, the algorithm calculates the token saving ratio (line 15). These patterns are then stored according to their token saving ratio. Subsequently, the algorithm stores each pattern except for those exceeding the savings threshold $T_{s}$ or falling below the elimination threshold $T_{e}$ in optimized semantic patterns (lines 19 and 20). The meta pattern information $MP$ is stored for future rounds. 

The algorithm proceeds to the next rounds and extends selected patterns in an iterative process. In each round, it evaluates patterns for their token saving ratio, excluding those that exceed the savings threshold $T_{s}$ or falling below the proportion threshold $T_{e}$. This iterative process of clustering, selection, and elimination continues until no patterns meet the criteria for exclusion or the maximum number of rounds $R$ is completed. The time complexity of SE-HSC is $O(R\cdot K)$. 


\begin{figure*}[t]
\centering
\begin{subfigure}{.49\textwidth}
  \centering
  \includegraphics[width=1\linewidth]{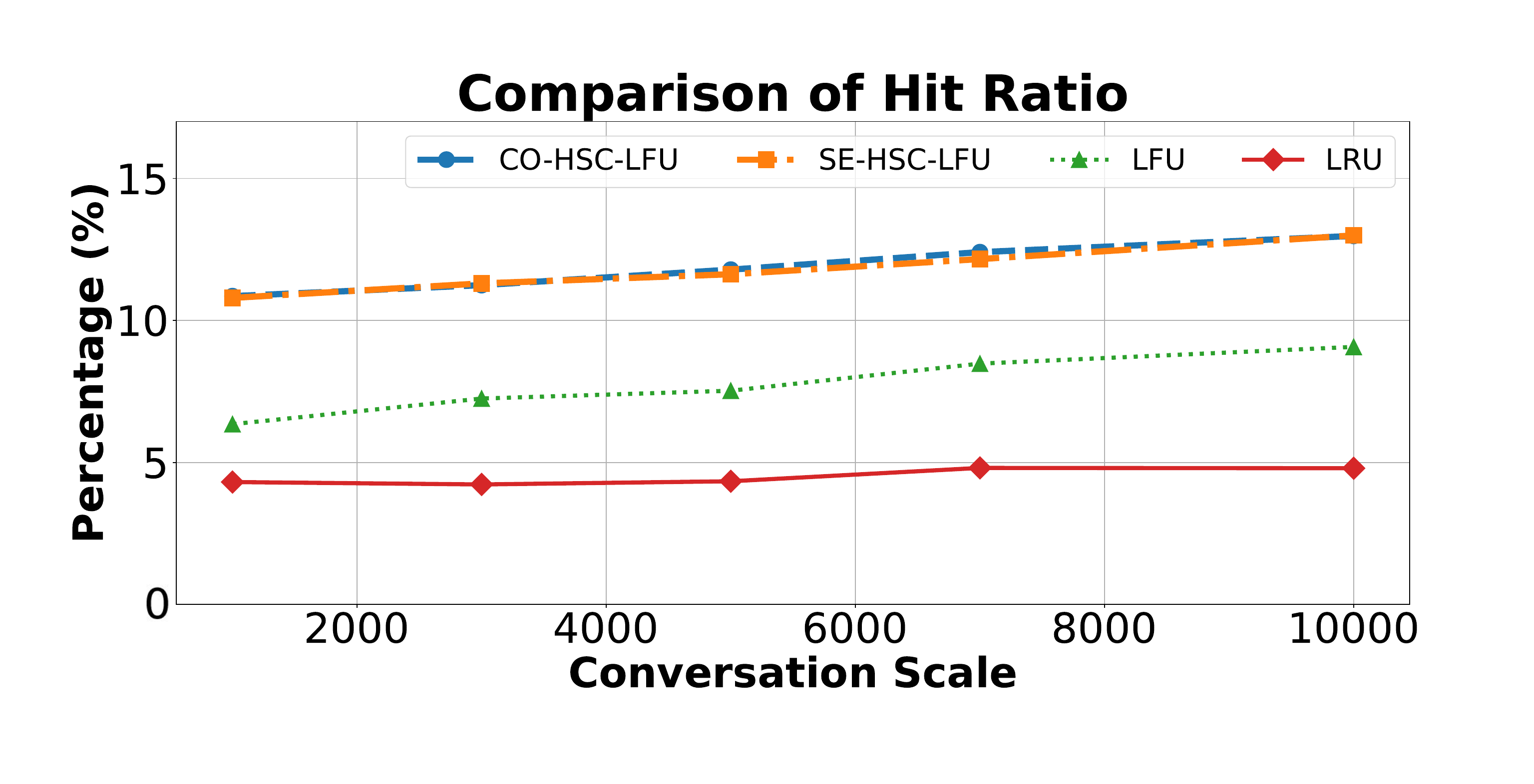}
\end{subfigure}
\begin{subfigure}{.49\textwidth}
  \centering
  \includegraphics[width=1\linewidth]{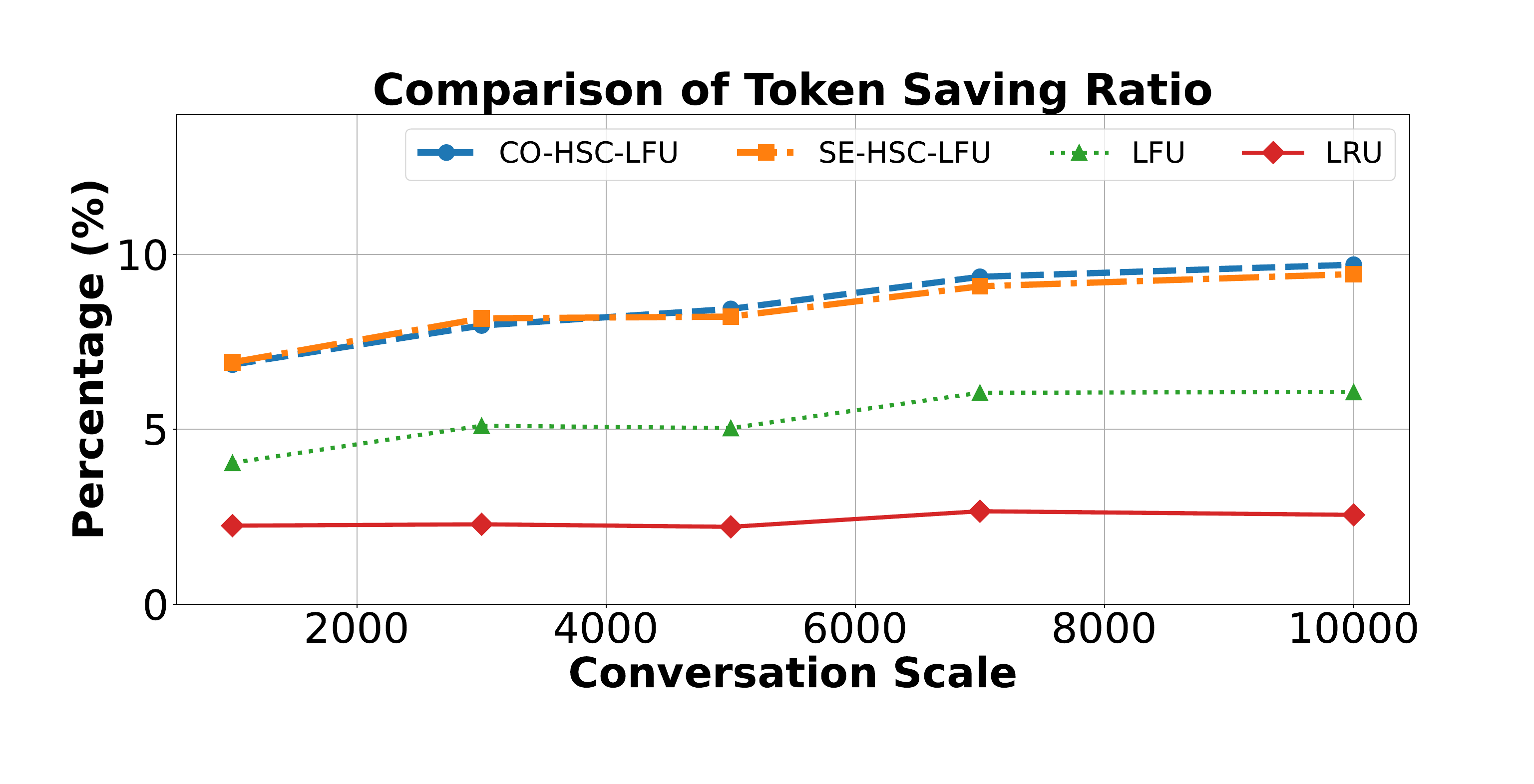}
\end{subfigure}%
\caption{Comparative analysis of hit ratios and token saving ratios with different conversation scales.}
\label{fig:conversation_scale}
\end{figure*}

\begin{figure*}[t]
\centering
\begin{subfigure}{.49\textwidth}
  \centering
  \includegraphics[width=1\linewidth]{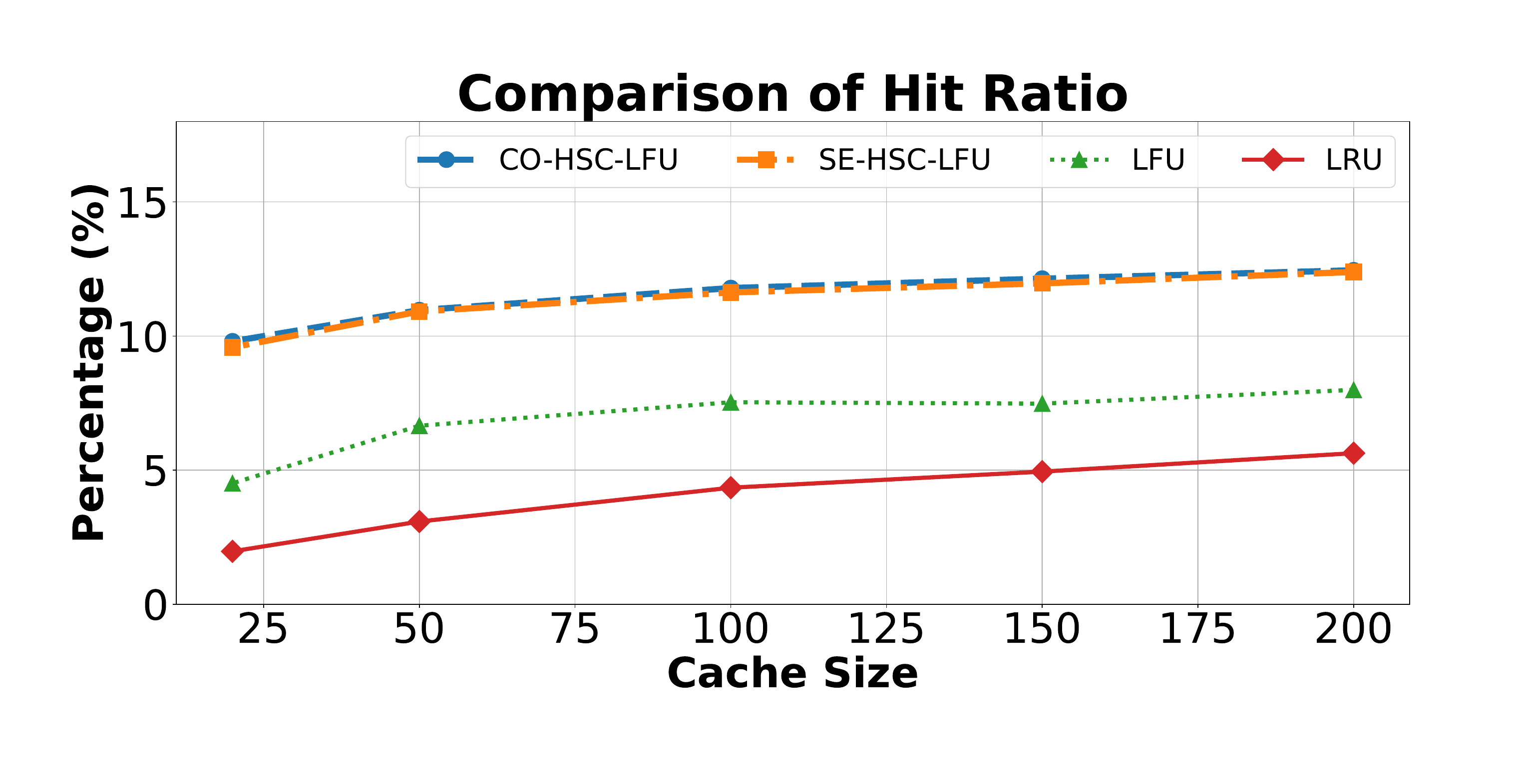}
  \vspace{-0.1cm}
\end{subfigure}
\vspace{-0.3cm}
\begin{subfigure}{.49\textwidth}
  \centering
  \includegraphics[width=1\linewidth]{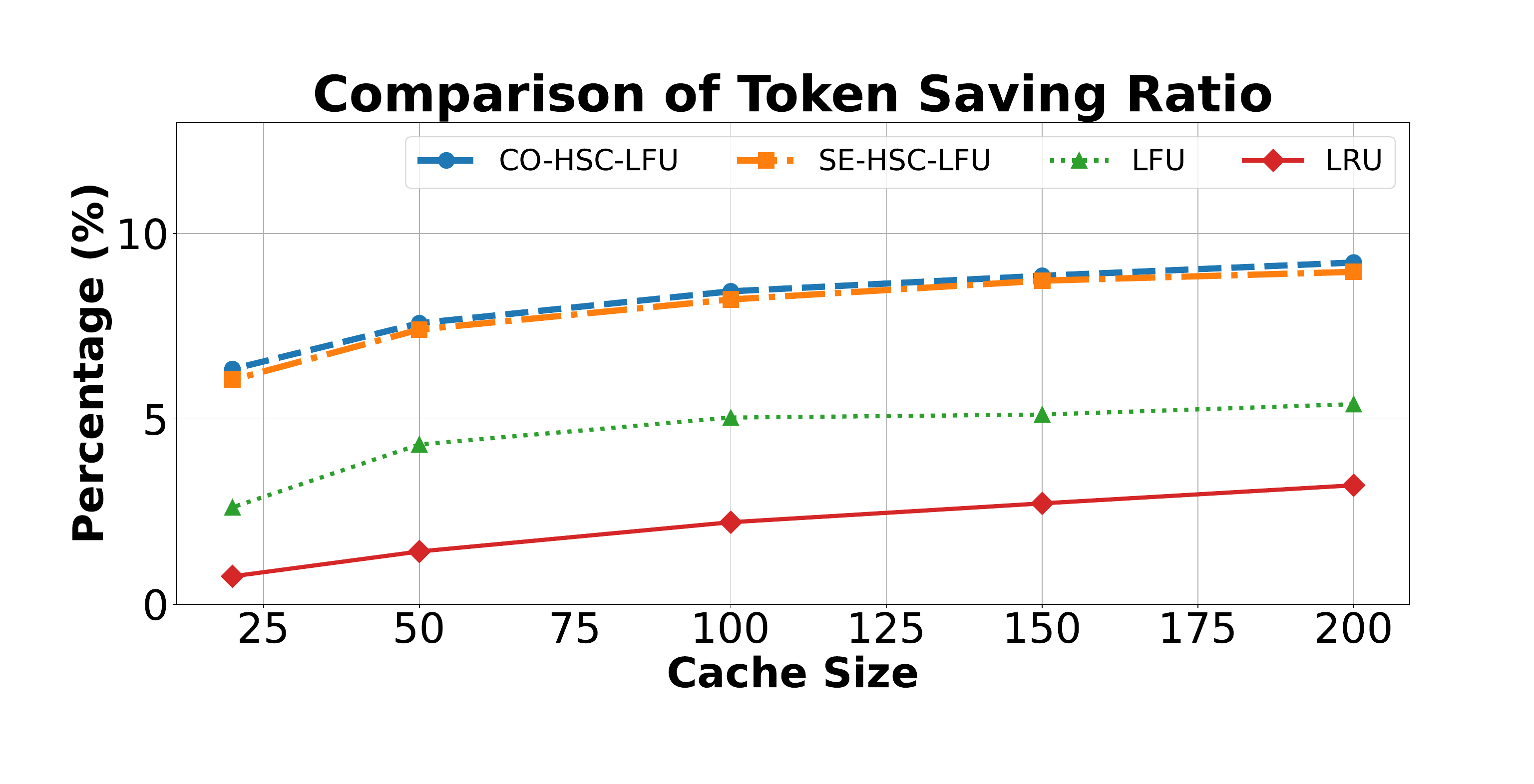}
  \vspace{-0.1cm}
\end{subfigure}%
\caption{Comparative analysis of hit ratios and token saving ratios with different cache sizes.}
\label{fig:cache_scale}
\end{figure*}

\section{Prototype Implementation and Evaluation}
\label{sec:implementation}
 
In this section, we present the implementation and evaluation of the SCALM prototype. We begin by introducing the clustering method and parameter settings for the CO-HSC and SE-HSC algorithms.


\subsection{Method and Parameter Settings}




In our exploration of different semantic clustering techniques, we consider DBSCAN~\cite{schubert2017dbscan}, k-means~\cite{ahmad2007kmean}, and other grid-based or density-based approaches~\cite{murtagh2012algorithms}. We find that DBSCAN shows superior performance, while k-means offers the fastest inference time but suffers from inaccuracy issues. Therefore, we choose the DBSCAN, one of density-based clustering method, for our prototype implementation.

We test different threshold settings for token savings and the proportion of patterns when employing the SE-HSC algorithm. To achieve the best performance with the SE-HSC algorithm, we suggest a criterion that patterns in each round achieve 20\% of the token saving ratio and constitute over 5\% of the dataset's weight to be considered for further extension and analysis in the next rounds. This criterion can be further adjusted to fit varying requirements or contexts. Similar to~\cite{zheng2023lmsys}, we initially categorize conversations into 20 distinct clusters to maintain efficiency and comprehensive coverage of conversation types. This number of clusters can be dynamically adjusted based on the characteristics of real-world usage.

We also define high, mid, and low ranks to prioritize certain semantic patterns. The high, mid, and low ranks label the top 25\%, 50\%, and 75\% of optimized semantic patterns, respectively. This ranking system helps inform caching and eviction strategies, allowing for the dynamic allocation of storage towards more significant patterns. Each rank's threshold is adaptable, offering flexibility to specific preferences or operational needs.

\subsection{Adaptive Storage Strategy}
\label{sec:storage_strategy}

We use hierarchical semantic clustering to optimize semantic patterns and guide our cache storage strategy. By considering the cache state and query pattern ranks, we adaptively update the storage threshold. The process starts by transforming incoming queries into vectors using the embedding model $E$, which then categorizes the embedding vectors into existing patterns. Depending on the cache state, either cold or full, we store queries based on their pattern ranks. Initially, low-rank patterns are stored during a cold start. As the cache fills, only queries from mid-rank and high-rank patterns are stored, tightening the criteria to enhance efficiency. This selective storage ensures that during peak times, only high-priority queries are retained, focusing on potential token savings and optimizing cache use.

\subsection{Adaptive Eviction Strategy}
\label{sec:eviction_strategy}

Our cache eviction strategy improves upon the LFU method by using optimized semantic patterns to evaluate the importance and frequency of queries. The eviction strategy is presented as follows.
When a new query arrives, the system classifies it into predefined semantic patterns and assigns an initial eviction priority value:
\begin{itemize}
    \item High-rank pattern queries start with a value of 3.
    \item Mid-rank pattern queries start with a value of 2.
    \item All other queries receive a value of 1.
\end{itemize}
Cache hits also increase the value. In this way, the cache eviction strategy ensures that more significant queries are less likely to be evicted early. Once the cache is full, the strategy removes the query with the lowest value. In the event of a tie, an older query is evicted.

\subsection{Metrics}
In evaluating the cache performance, the most commonly used metric is hit ratio $R_{hit}$, which is defined as follows:

\begin{equation}
R_{hit} = \frac{N(\textbf{\text{H}})}{N(\textbf{\text{Q}})}
\end{equation}
where $\textbf{\text{H}}$ is the set of cache hits and $\textbf{\text{Q}}$ is the set of queries. $N(\cdot)$ denotes the count of the set entries. Since the cache is designed for LLMChat services to reduce the number of tokens processed, we propose another important performance metric: $R_{ts}$, the token saving ratio brought by the successful cache hits. Intuitively, the count of tokens processed for a query $q$ is calculated by summing up the count of tokens in the user query and LLMChat response. Additionally, it is worth noting that the context or the previous queries and answers are also processed when LLMs handle the current query. Formally, token saving ratio $R_{ts}$ can be calculated as follows: 

\begin{equation}
R_{ts} = \frac{\sum\limits_{q \in \textbf{\text{H}}}{C(q)}}{\sum\limits_{q \in \textbf{\text{Q}}}{C(q)}}
\end{equation}
where $C(\cdot)$ denotes the count of processed tokens by LLMs.



\subsection{Performance Evaluation}

We develop a SCALM prototype by integrating semantic analysis algorithms into GPTCache, referencing the base LFU cache replacement algorithm. Consequently, we label our methods as CO-HSC-LFU and SE-HSC-LFU and use LFU and LRU as baselines in GPTCache. Through adjustments to cache size and conversation scale, we observe performance improvements in real-world experiments: the hit ratio increased by approximately 5.5\% and the token saving ratio by around 4.6\%. Considering that the current optimal cache performance on content distribution is around 20\%, these results achieved by SCALM are significant.



Figure~\ref{fig:conversation_scale} shows the performance of CO-HSC-LFU, SE-HSC-LFU, LFU, and LRU across different experimental scales, ranging from 1,000 to 10,000 conversations, with five representative cases selected. The figure clearly shows that both of our strategies, CO-HSC-LFU and SE-HSC-LFU, outperform the baseline methods across all conversation scales. For example, with a conversation scale of 5000, the cache hit ratios of CO-HS-LFU and SE-HSC-LFU are 11.7\% and 11.6\%, respectively, surpassing the baseline LFU by over 4.3\%. The token saving ratios in this case are 8.4\% and 8.2\% for CO-HSC-LFU and SE-HSC-LFU, respectively. Typically, CO-HSC-LFU performs slightly better than SE-HSC-LFU. This is expected as CO-HSC-LFU retains all clusters in semantic analysis, whereas SE-HSC-LFU uses strategic pruning to improve computational resources by focusing on key clusters, which sometimes results in a performance trade-off.




In Figure~\ref{fig:cache_scale}, we show the performance of CO-HSC-LFU, SE-HSC-LFU, LFU, and LRU across cache sizes ranging from 20 to 200, with five representative cases shown. The improvements from CO-HSC-LFU and SE-HSC-LFU are stable across different cache sizes. The results reveal that neither conversation scale nor cache size impacts the effectiveness of CO-HSC-LFU and SE-HSC-LFU. Both algorithms consistently outperform the baseline in hit ratio and token saving ratio, demonstrating their efficiency and robustness.



\begin{figure}[t]
    \centering    \includegraphics[width=0.5\textwidth]{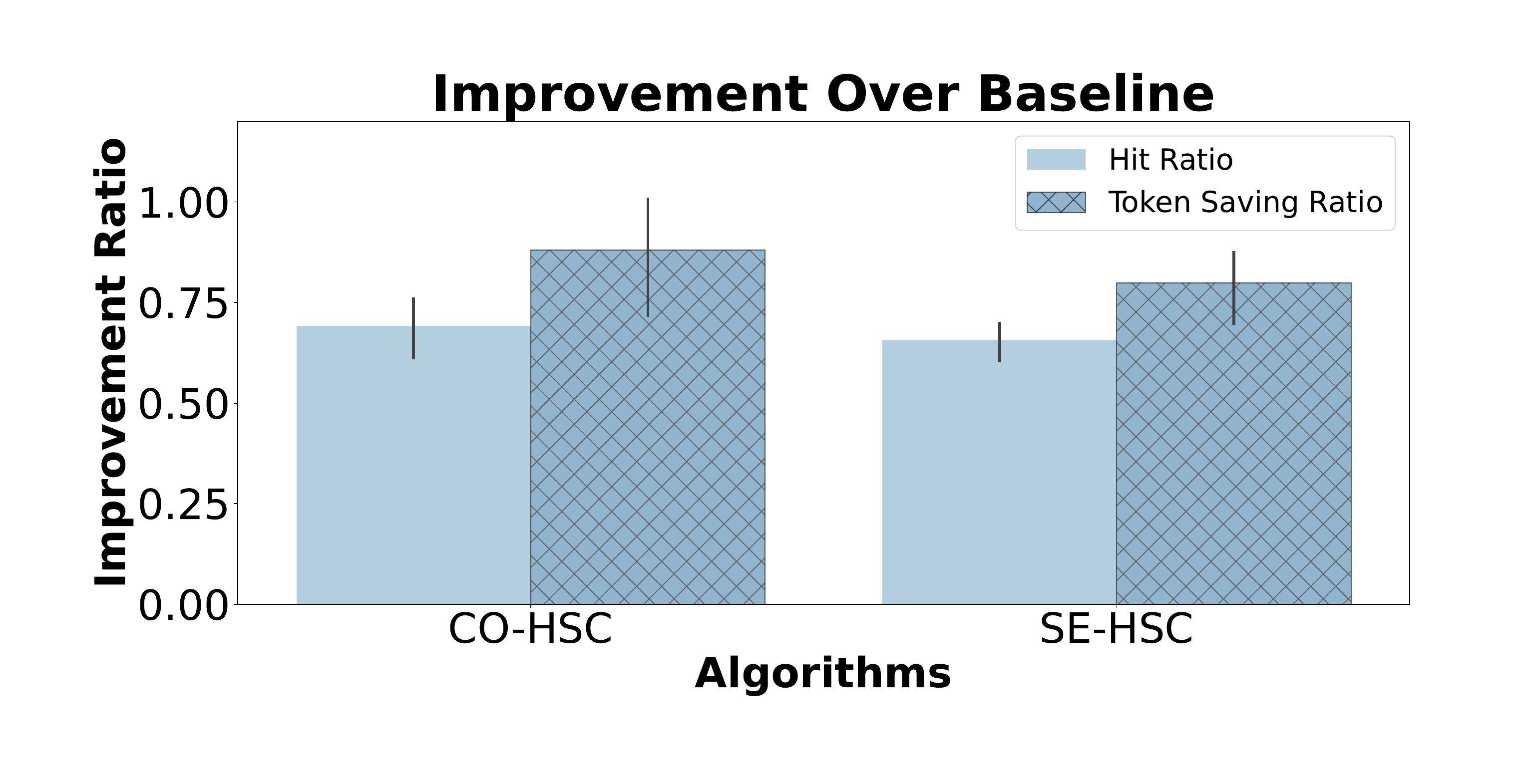}
    \vspace{-0.5cm}
    \caption{Relative improvement ratio compared to LFU.}
    \label{fig:improvement_ratio}
\end{figure}

In Figure \ref{fig:improvement_ratio}, we present the relative enhancement ratio of CO-HSC-LFU and SE-HSC-LFU over the baseline LFU. In this experiment, we fix the conversation scale at 3,000 and the cache size at 100 and perform the experiments 10 times. This figure demonstrates that, on average, the hit ratio can achieve approximately a 63\% relative enhancement, while the token saving ratio exhibits a relative enhancement of around 77\%. 

\begin{figure}[t]
    \centering
    \begin{subfigure}{0.5\textwidth}
        \centering
        \includegraphics[width=\linewidth]{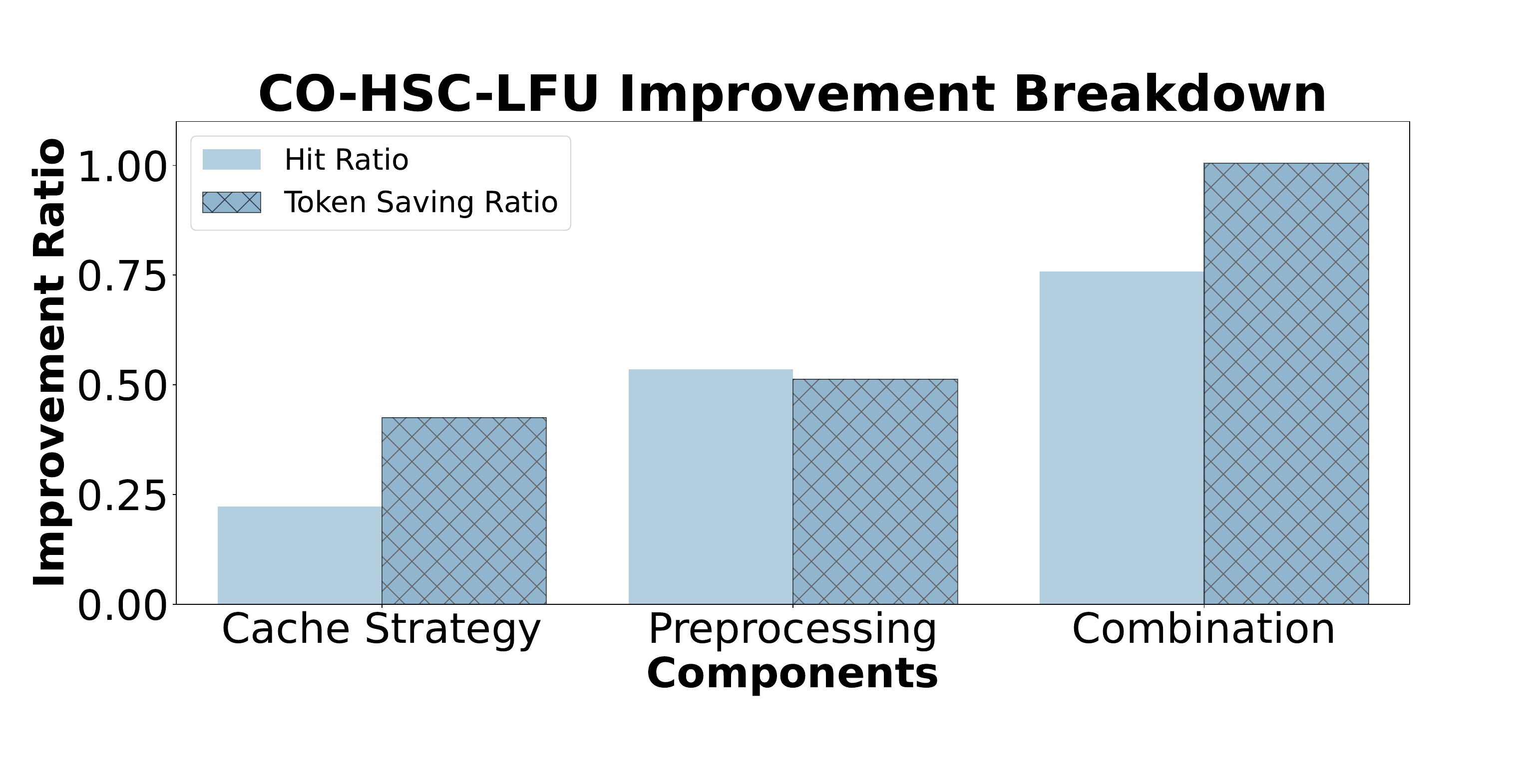}
        \vspace{-0.5cm}
        \label{fig:improvement_ratio_CO-HSC_sub}
    \end{subfigure}
    \vspace{-0.5cm}
    \begin{subfigure}{0.5\textwidth}
        \centering
        \includegraphics[width=\linewidth]{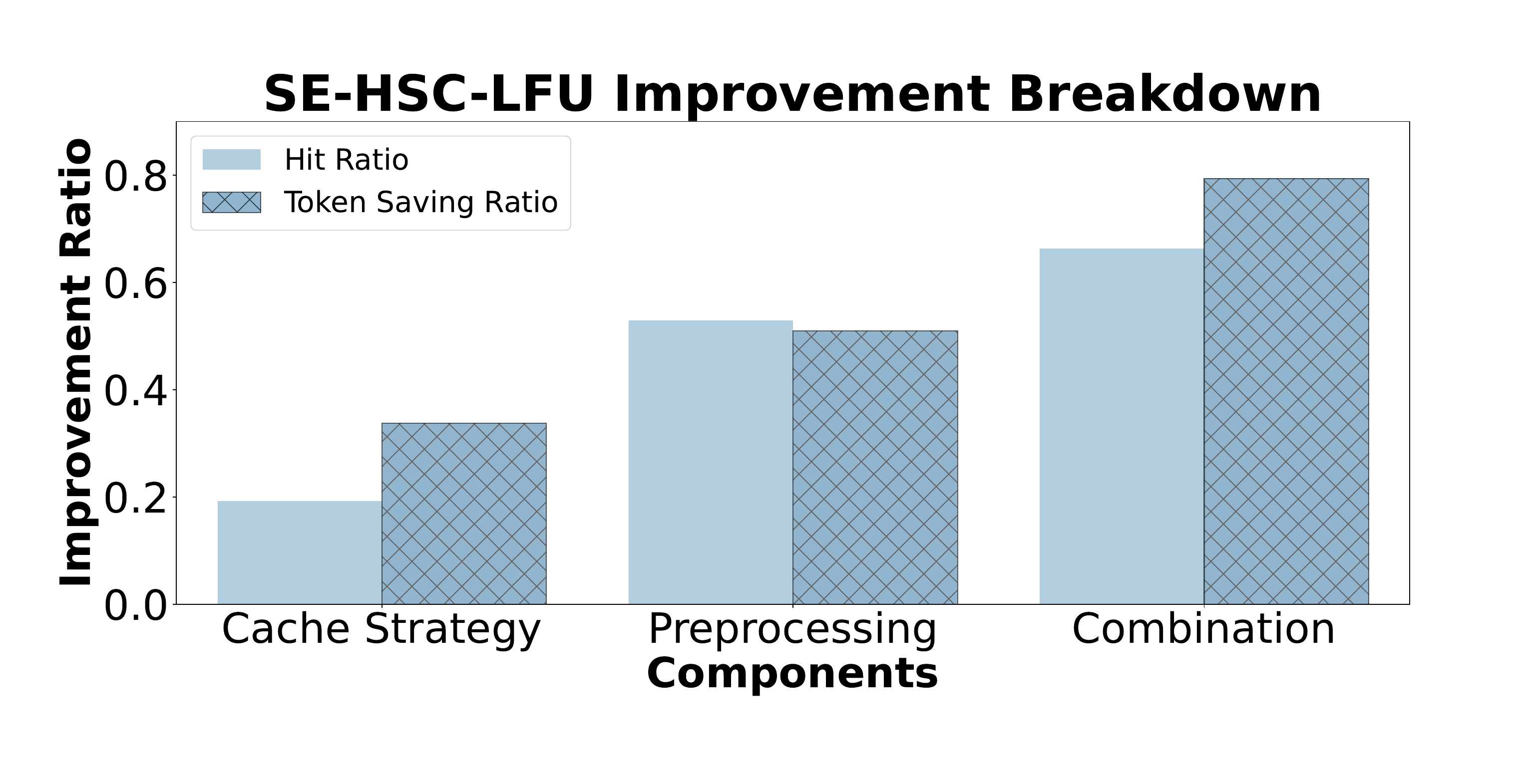}
        \label{fig:improvement_ratio_SE-HSC_sub}
    \end{subfigure}
    \vspace{-0.3cm}
    \caption{Breakdown of improvement ratios.}   \label{fig:improvement_ratios_combined}
\end{figure}

We conduct ablation experiments to assess the contributions of components in the SCALM architecture. Specifically, we analyze the impact of cache strategies and enhanced query preprocessing (in similarity evaluation) on CO-HSC-LFU and SE-HSC-LFU individually. Figure~\ref{fig:improvement_ratios_combined} illustrates the results, indicating that both cache strategies and enhanced preprocessing contribute to the improvements. Cache strategies alone result in relative token-saving improvements ranging from 35.2\% to 38.1\%, while enhanced preprocessing yields enhancements between 46.2\% and 48.2\%, both compared to the baseline LFU. Furthermore, the combination of these components amplifies the enhancements, highlighting SCALM's focus on token efficiency.

\section{Further Discussion}
\label{sec:future}

\textbf{Sub-optimal Cache Performance.} Our insights and design are based on empirical investigations, suggesting the potential for further refinement. Moving forward, we aim to incorporate semantic analysis into the online optimal cache algorithm. This optimality arises from analyzing the underlying query distributions. Belady’s offline optimal algorithm~\cite{belady1966study} serves as an upper bound of all cache algorithms. Recent advancements in learning-based cache algorithms~\cite{liu2020imitation} leverage neural networks to learn control policies from Belady’s. Integrating such a learning-based algorithm could enhance the robustness of our cache design.

\textbf{Multimodal Response Caching.}
Modern LLMs like GPT-4 now support image and video generation~\cite{rombach2022highresolution,ho2022imagenvideo,wu2022nuwa,openai2023videogeneration}. In some cases, it might take several minutes to generate a short video clip. For example, we utilize Runway Gen2\footnote{\url{https://research.runwayml.com/gen2}} to generate videos from the text. Gen2 represents one of the top-tier models in today's text-to-video generation landscape \cite{dirik2023texttovideo}. We found that a 4-second video took more than 90 seconds. This extended generation time can often impact user experience, potentially leading to reduced satisfaction. Recognizing this, the concept of caching multimodal response emerges as a potential solution. 
Similarly, when serving multimodal queries, it is also crucial to reconsider the design and implementation of its caching mechanism. 

\textbf{Extension to Semantic Graphs.} We will further explore semantic graphs as an alternative to cache storage design~\cite{zhu2023semantics}. In a semantic graph, nodes represent dialogue entries, while edges denote extended similarity relationships. Each node and edge carries attributes that detail its characteristics and relationships. This graph storage enables the application of subgraph matching algorithms to efficiently identify patterns or query-specific substructures. Transforming dialogues into a semantic graph will facilitate a more flexible search mechanism. It not only enhances the accuracy of matching queries but also supports fuzzy search capabilities.

\textbf{Scaling the Semantic Cache.} As LLMs serve increasingly complex and numerous user queries \cite{LLMNetworking}, the semantic cache will be distributed across multiple machines to manage the load effectively. This requires the use of distributed caching technologies that support fast data retrieval and high availability~\cite{xu2023sparksgpt}. 

Although our implementation primarily focuses on single-machine settings, it is extendable to distributed settings. Techniques such as consistent hashing~\cite{bang2023gptcache} can be employed to evenly distribute cache data across machines, ensuring that the system can scale horizontally as demand grows~\cite{xu2018enhancing}. Regarding hash key design, organizing data around similar semantic patterns allows queries related to a specific pattern to be served by the same machine or a smaller subset of machines. It is straightforward to adapt the semantic patterns extracted by our hierarchical semantic clustering to serve as hash keys. Similar Q\&A entries belonging to the same semantic pattern will be stored on the same machine, thereby localizing data access.



\section{Conclusion}
\label{sec:conclusion}
In this work, we conducted real-world data-driven analyses on the cache performance of automated chat services with LLMs. Based on our findings, we proposed a new cache architecture named SCALM. This architecture employs semantic analysis to identify significant cache entries and their semantic patterns. When a new query arrives, it undergoes extensive comparisons with the semantic patterns, informing storage and eviction decisions. Performance assessments indicate that SCALM improves cache efficiency and diminishes operational expenditures for LLMChat services. On average, there is a 63\% augmentation in cache hit ratio and a 77\% reduction in token consumption when compared to the established benchmarks of the cutting-edge GPTCache framework.


\section*{ACKNOWLEDGEMENTS}
This research was partly supported by an NSERC Discovery Grant. The corresponding authors of this paper are Cong Zhang and Jiangchuan Liu.

\let\oldbibliography\thebibliography
\renewcommand{\thebibliography}[1]{
  \oldbibliography{#1}
  \footnotesize  
  \setlength{\itemsep}{0pt}
}
\balance
\bibliographystyle{IEEEtran}
\bibliography{reference}

\end{document}